\documentclass[nohyperref]{article}

\usepackage{microtype}
\usepackage{graphicx}
\usepackage{subfigure}
\usepackage{booktabs} 
\usepackage{adjustbox}
\usepackage{hyperref}

\usepackage[table, dvipsnames]{xcolor}

\usepackage[accepted]{icml2023}

\usepackage{amsmath}
\usepackage{amssymb}
\usepackage{mathtools}
\usepackage{amsthm}

\usepackage[capitalize,noabbrev]{cleveref}

\theoremstyle{plain}

\theoremstyle{definition}

\theoremstyle{remark}

\usepackage[textsize=tiny]{todonotes}

\icmltitlerunning{Pruning for Better Domain Generalizability}

\begin{document}

\twocolumn[
\icmltitle{Pruning for Better Domain Generalizability}




\begin{icmlauthorlist}
\icmlauthor{Xinglong Sun}{yyy}
\end{icmlauthorlist}

\icmlaffiliation{yyy}{Department of Computer Science, Stanford University}

\icmlcorrespondingauthor{Xinglong Sun}{xs15@stanford.edu}

\icmlkeywords{Machine Learning, ICML}

\vskip 0.3in
]



\printAffiliationsAndNotice{}  

\begin{abstract}
In this paper, we investigate whether we could use pruning as a reliable method to boost the generalization ability of the model. We found that existing pruning method like L2 can already offer small improvement on the target domain performance. We further propose a novel pruning scoring method, called DSS, designed not to maintain source accuracy as typical pruning work, but to directly enhance the robustness of the model. We conduct empirical experiments to validate our method and demonstrate that it can be even combined with state-of-the-art generalization work like MIRO\cite{cha2022domain} to further boost the performance. On MNIST to MNIST-M, we could improve the baseline performance by over $5$ points by introducing $60\%$ channel sparsity into the model. On DomainBed benchmark and state-of-the-art MIRO, we can further boost its performance by 1 point only by introducing $10\%$ sparsity into the model. Code can be found at: https://github.com/AlexSunNik/Pruning-for-Better-Domain-Generalizability. 
\end{abstract}
\section{Introduction}
\label{sec:intro}
In recent years, we have seen rapid development in deep neural networks with various model architectures\cite{he2016deep, dosovitskiy2020image} for a wide range of different tasks. However, the pretrained models could be poor at generalizing learned features or knowledge to new datasets or environments. Even a slight shift from the network's original training domain could significantly hurt its performance\cite{recht2019imagenet, hendrycksbenchmarking, yang2021generalized}, which suggests that the successes achieved by deep learning so far have been largely driven by supervised learning and large-scale labeled datasets\cite{deng2009imagenet}. This domain shift issue has seriously impeded the development and practical deployment of deep neural models. \par
A straightforward approach to deal with this domain shift is to collect some data from the target domain to adapt the source-domain trained model, relying on the assumption that target data is accessible for model adaption. This \emph{domain adaptation} approach\cite{lu2020stochastic, saito2018maximum, ganin2015unsupervised, long2015learning, liu2020open, hoffman2018cycada, gong2012geodesic, long2016unsupervised, balaji2019normalized, kang2019contrastive, kulis2011you, gandelsman2022test, sun2020test, liu2021ttt++} is practical in many scenarios since we do not need extra labor and computation cost to label the collected target distribution data and simply leverage the unlabeled data for adaptation. However, it does assume that we have access to target domain data during training, which is infeasible and impractical in many cases.\par
A more strictly defined problem is \emph{domain generalization}, which does not assume access to target sample features during training and strives to learn robust representations against distribution shifts from multiple source domains during training. We later perform evaluation of the trained model on an unseen test domain to measure the generalizability, transferability, and robustness of the model. While many existing works in domain generalization attempt to learn domain-invariant features~\cite{arjovsky2019invariant, bui2021exploiting, cha2021swad, ganin2015unsupervised, li2018learning, sun2016deep}, some recent works~\cite{gulrajanisearch, koh2021wilds} also demonstrate decent accuracies without explicitly enforcing invariance. Recently, the state-of-the-art work MIRO~\cite{cha2022domain} aims to learn similar features to 'oracle' representations, reformulating the domain generalization task by maximizing mutual information between oracle representations and model representations while performing training on source domains. It achieves state-of-art performance on the widely-used DomainBed~\cite{gulrajanisearch} benchmark.\par 
A seemingly completely irrelevant direction, pruning~\cite{lecun1989optimal,hassibi1993optimal,journals/corr/HanPTD15,frankle2018the,LeeAT19, sun2022disparse}, aims to compress the model by removing the least important channels scored by some saliency criterion. Since size of the model gets shrinked after pruning, it is also sometimes considered to reduce the overfitting of models and increase the domain generalization ability of the model from another perspective. In fact, some latest works~\cite{li2022can, jin2022pruning, bartoldson2020generalization} began to investigate deeper relationship between pruning and the generalizability and robustness of models. \par

In this paper, we made a further step, investigating whether we could use pruning as a reliable method to boost the generalization ability of the model. We aim to answer the following three questions:
\begin{enumerate}
    \item Can we leverage existing popular and simple pruning metrics like L2\cite{conf/iclr/0022KDSG17} to boost generalization accuracy by pruning unimportant channels? 
    \item Can we design a better pruning score taking the generalization ability of the model into consideration? More concretely, a score specifically designed to improve target domain accuracy instead of maintaining source domain accuracy as typical pruning.
    \item  Finally, can we combine it with modern state-of-the-art domain generalization algorithms like MIRO\cite{cha2022domain} as a simple plug-in component to further boost the accuracy?
\end{enumerate}
We answered the above three questions with solid empirical studies ranging across three datasets and model architectures. To begin with, we study the first two questions extensively across many different pruning sparsity ratios on MNIST to MNIST-M, which is randomly colored MNIST. We found that the existing simple pruning method L2\cite{conf/iclr/0022KDSG17} can offer a small improvement over the vanilla baseline(i.e. without using any domain generalization technique). Later, we solve the question (2) by designing a novel pruning method specifically targeting generalization accuracy. Given a convolutional neural networks(CNNs), we evaluate the activation map for samples from different domains at each layer for every channel and compute a \emph{domain similarity score (DSS)} based on the distance of the activation maps. We then use \emph{structural pruning} to prune the channels with the lowest DSS, followed by a finetuning session to recover accuracy. From empirical results, we observe an obvious improvement from the standard L2 pruning score. Notably, we can improve the baseline performance by more than $5$ points by sparsifying $60\%$ of the channels in the model, which may seem very surprising.\par
After validating the effectiveness of our proposed DSS score, we resolve the question (3) by combining our method with the state-of-the-art work MIRO~\cite{cha2022domain}. We conduct experiments on two datasets PACS and OfficeHome from the DomainBed~\cite{gulrajanisearch} benchmark and observe a $1$ point improvement of MIRO by introducing a $10\%$ channel sparsity into the model, demonstrating the capability of our method to even improve the SOTA result.

\section{Related Works}
\subsection{Domain Generalization}
Many works aim to tackle domain generalization task through learning domain-invariant features by either minimizing between-domain feature divergences~\cite{ganin2015unsupervised, li2019episodic, matsuura2020domain, sun2016deep, zhao2020domain}, robust optimization~\cite{arjovsky2019invariant, cha2021swad}, or augmenting source domain examples~\cite{bai2021decaug, carlucci2019domain}. Inspired by prior works, our pruning DSS score also aims to remove features that are the most domain-sensitive, or in other words, keep the most domain-similar features.

\subsection{Pruning}
Network pruning methods can be roughly categorized as \emph{unstructured pruning} and \emph{structured pruning}.
Unstructured pruning methods~\cite{lecun1989optimal,hassibi1993optimal,journals/corr/HanPTD15,frankle2018the,conf/iclr/LeeAT19} removes individual neurons of less importance without consideration for where they occur.
On the other hand, structured pruning methods~\cite{conf/iclr/0022KDSG17,liu2017learning,conf/iclr/MolchanovTKAK17} prune parameters under structure constraints, for example removing convolutional filters. Most pruning methods focus on designing an importance score to reflect parameters' importance to the final output. The popular channel pruning score L2~\cite{conf/iclr/0022KDSG17} is evaluated as the Frobenious Norm of the convolution kernels. In this work, we leverage structural pruning as well because it allows us to filter and select features. Unlike all previous pruning works which aim to shrink the model size as much as possible while maintaining good source performance, our method focuses on enhancing the generalization ability of the model. We design a novel score \emph{domain similarity score (DSS)} for this purpose, which measures the similarity of features between the source and target domain to keep robust features for estimating output.

\section{Methodology}
\begin{figure}[t!]
    \centering
    \includegraphics[width=.8\linewidth]{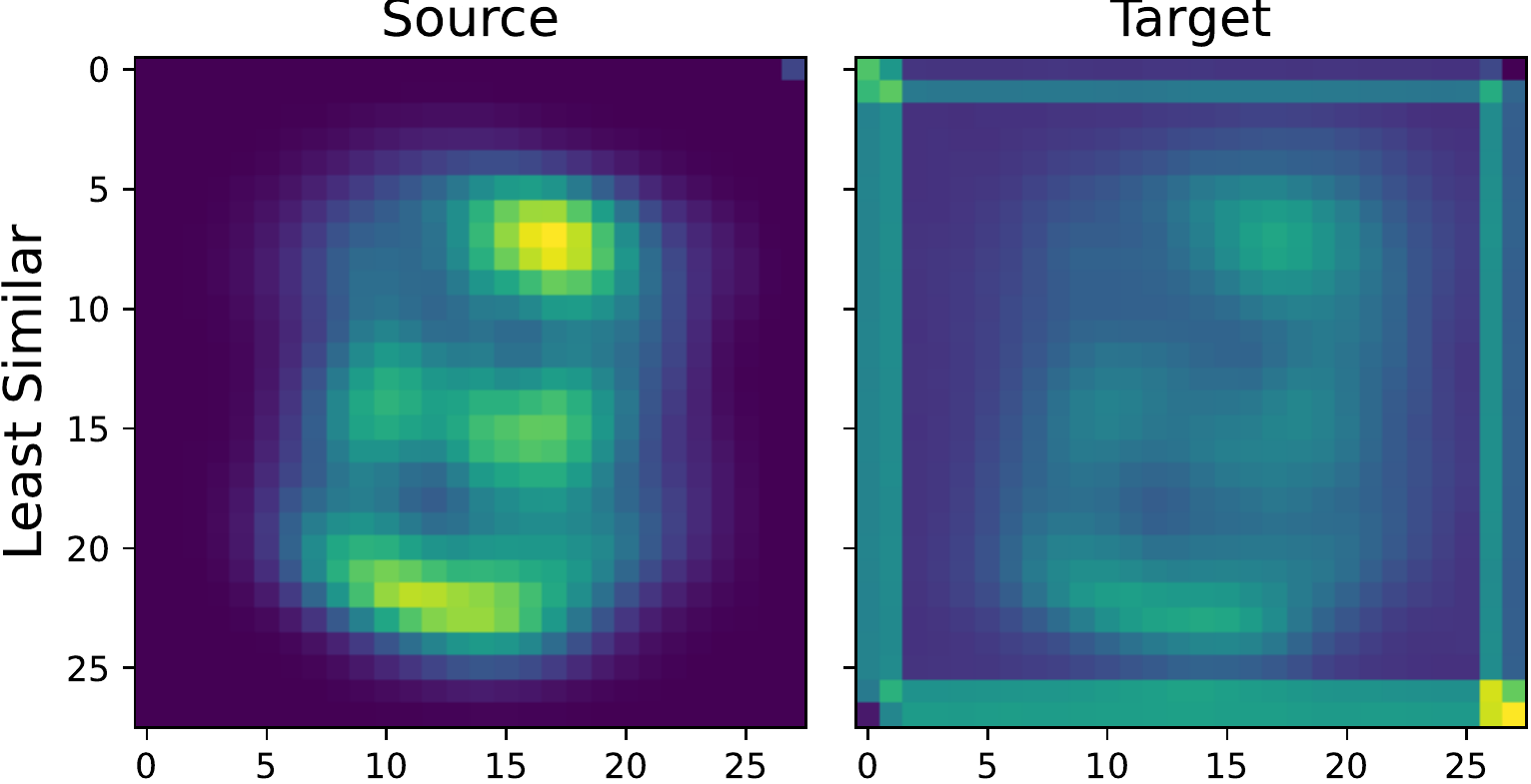}
    \includegraphics[width=.8\linewidth]{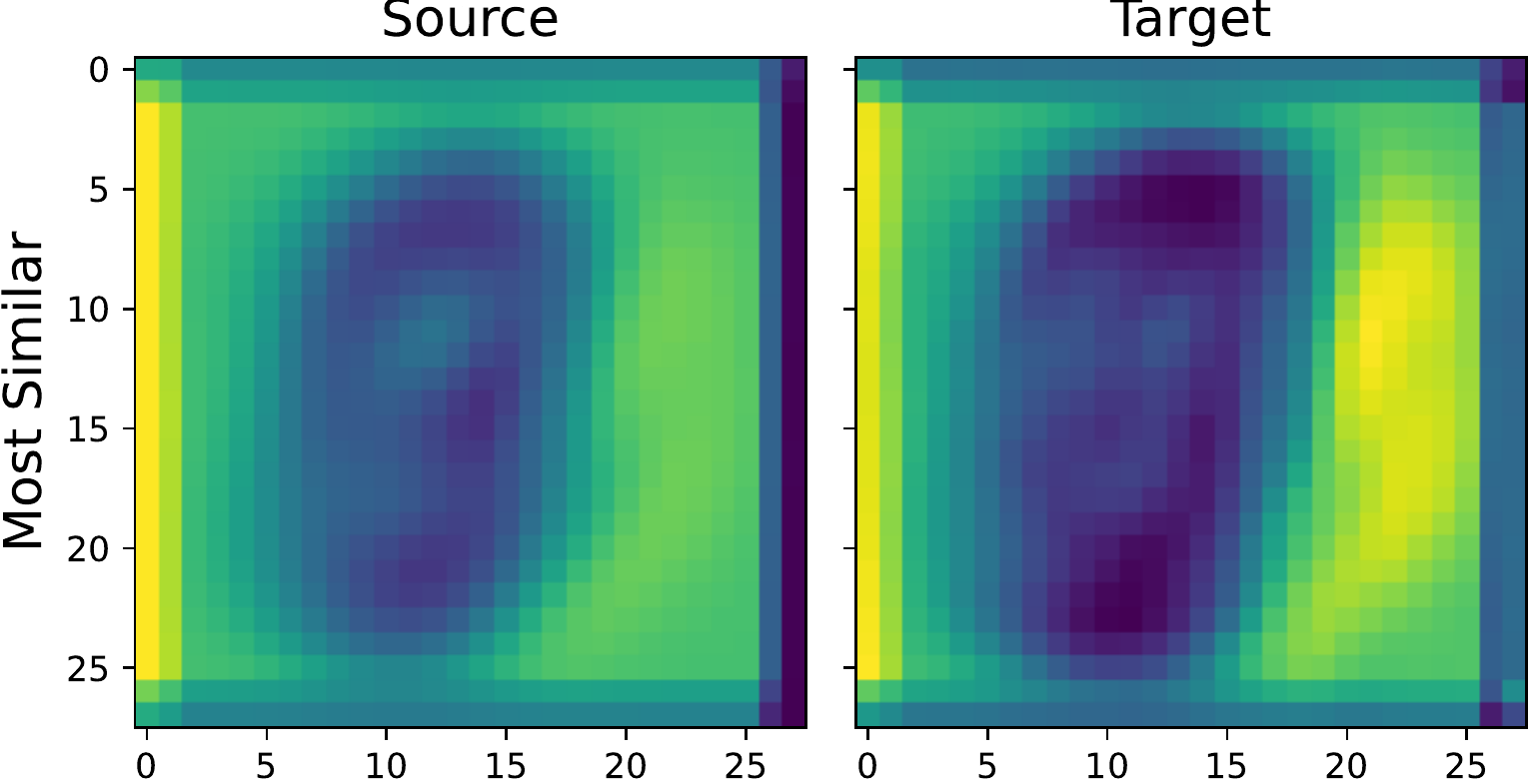}
    \caption{\textbf{MNIST to MNIST-M} Visualization of \emph{averaged} feature maps that are least and most similar between source and target domains ranked by our DSS score. In the least similar case, we could see that the target feature map contains highlighted regions in the corner, which suggests potential \emph{spurious features}.}
    \label{fig:feat_viz}
\end{figure}

Here, we describe how to compute the proposed DSS score. Consider $\mathcal{X}$ as the input space and $\mathcal{Y}$ as the output space. We define a domain as a joint distribution $P_{XY}$  on $\mathcal{X} \times \mathcal{Y}$. Moreover, we refer to $P_X$ as the marginal or input distribution on $X$, and $P_{Y|X}$ as the posterior distribution of $Y$ given $X$. Suppose we have two distinct domains, $P_{XY}^1$ and $P_{XY}^2$ ($P_{XY}^2 \neq P_{XY}^1$). \par
Suppose we are dealing with image data and suppose $x^1 \sim P_X^1, x^1 \in \mathbb{R}^{c\times h \times w}$, and $x^2 \sim P_X^2, x^2 \in \mathbb{R}^{c\times h \times w}$, here, $c, h, w$ are input channels(3 for RGB), height, and weight of our input data to the model. Below for convenience, when we do not distinguish the domain, we simply use $x$ to refer to the input image. In representation learning, we usually build a feature extractor $T(.)$ constructing representation from raw input data $x$ and outputting features maps or activation maps. The extracted features are then fed into a projector or linear classifier $f(.)$ to produce the final estimation, $i.e. f(T(x))$. Moreover, suppose $T(x) \in \mathbb{R}^{\hat{c} \times \hat{h} \times \hat{w}}$.\par
Given $x^1$ and $x^2$, we construct source and domain feature maps respectively as $T(x^1)$ and $T(x^2)$. The goal of our DSS score is to measure the similarity of $T(x^1)$ and $T(x^2)$. To this end, we simply leverage the score proposed in DeepFace~\cite{taigman2014deepface} which computes the normalized inner product of the flattened features. We denote normalize and flatten operator as $\gamma(.)$ for simplicity. Therefore, at a convolutional layer, {domain sensitivity score (DSS)} $\mathcal{S}$ for each channel $i$ out of $\hat{c}$ channels can be computed as:
\begin{align}
    \mathcal{S}_i &= \langle\mathbb{E}[\gamma(T(x^1)_i)], \mathbb{E}[\gamma(T(x^2)_i)]\rangle \\
    \mathcal{S} &= [\mathcal{S}_0, \mathcal{S}_1, \hdots, \mathcal{S}_{\hat{c}}]
\end{align}
In practice, we could leverage Monte Carlo Estimation for estimating the expected features from each distribution. For $\mathbb{E}[\gamma(T(x^1)_i)]$ for example, we can compute it as:
\begin{align}
    \mathbb{E}[\gamma(T(x^1)_i)] = \frac{1}{N}\sum_{x^1 \sim P_X^1}^{N}\gamma(T(x^1)_i),
\end{align}
which samples $N$ times from the distribution $P_X^1$. A similar procedure can be done for $P_{XY}^2$ as well.
With the computed DSS score, we then prune the channels given by $\text{ArgBotK}(\mathcal{S}, n)$, which selects the bottom $n$ channels out of $\hat{c}$ channels with the lowest DSS. We finish by performing a finetuning session on the kept channels to recover accuracy.\par
In Figure.\ref{fig:feat_viz}, we demonstrate the example feature maps on MNIST-M dataset with lowest(most sensitive) and highest (most similar) feature maps respectively. Interestingly, in the least similar case, we could see that the target feature map contains very highlighted regions in the corner of the image, which suggests potential \emph{spurious features} since MNIST-like dataset mostly does not contain \emph{'useful'} features there.

\section{Empirical Results}
As mentioned in Section\ref{sec:intro}, we conduct experiments with the augmented MNIST dataset MNIST-M and two datasets PACS and OfficeHome from DomainBed~\cite{gulrajanisearch}. MNIST-M dataset contains digits from original MNIST dataset blended over patches randomly extracted from color photos of BSDS500. Since both datasets are relatively not large in size, in selecting $N$(number of times to estimate expected feature maps), we just leverage the entire dataset. Moreover, the methodology section only discusses procedure on a single layer. With a multi-layered network like ResNet~\cite{he2016deep}, we perform the described procedure at every layer independently.

\begin{figure}[t!]
    \centering
    \includegraphics[width=\linewidth]{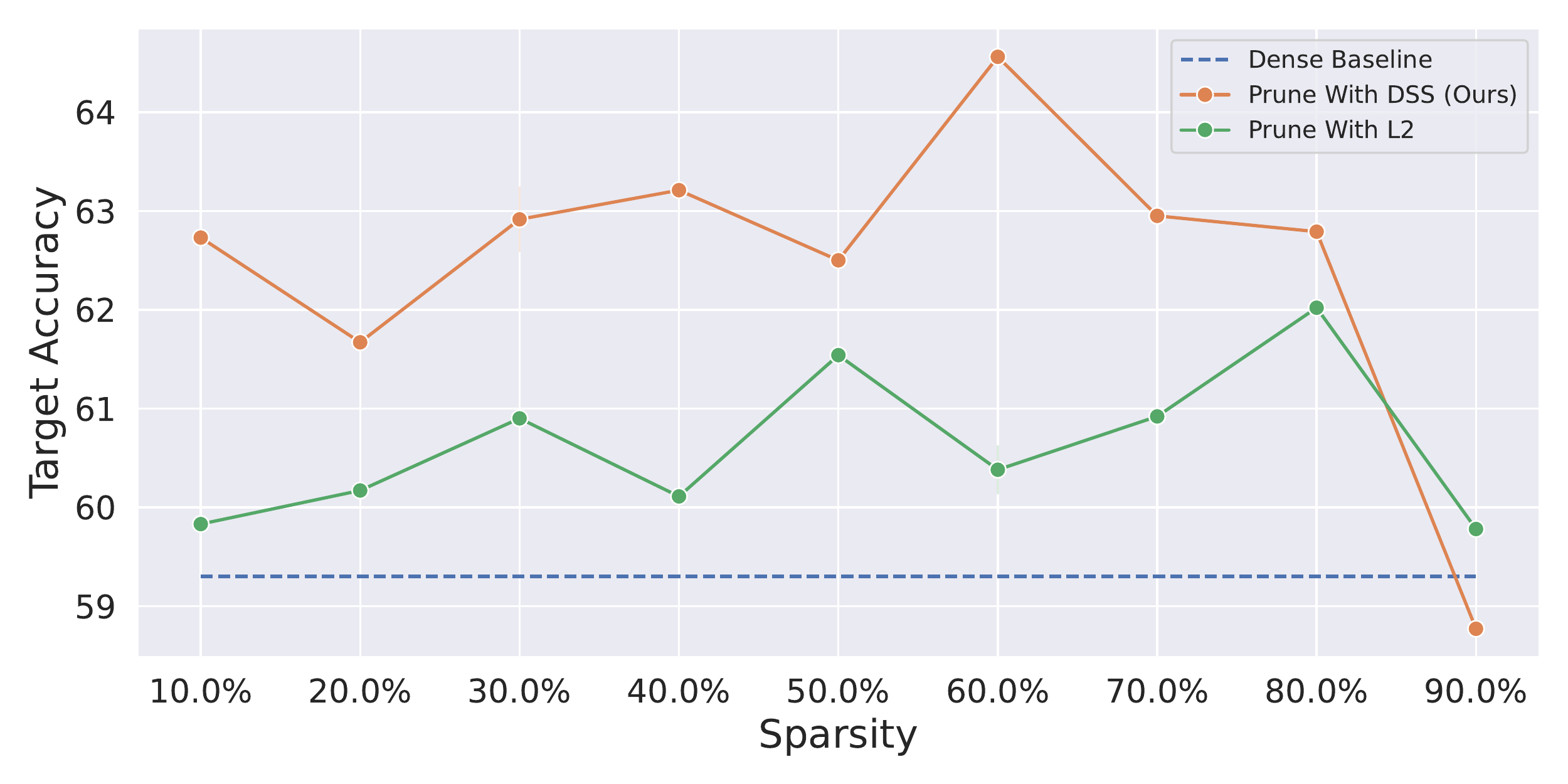}
    \vspace{-15pt}
    \caption{\textbf{MNIST to MNIST-M} Performance at different channel sparsity ratios. Our DSS scores reliably improve target accuracy and performs better than standard pruning scores like L2~\cite{conf/iclr/0022KDSG17}.}
    \label{fig:mnist_res}
    \vspace{-15pt}
\end{figure}

\begin{table}[t!]
    \centering
    \begin{tabular}{c|cc}
    \toprule
     Method & Before Ft. & After Ft.\\ \midrule
     Baseline  &  \multicolumn{2}{c}{59.30} \\
     Rev.DSS  & 55.93 & 58.35\\
     \rowcolor{lightgray}
     \textbf{DSS} & 61.14  & 63.21\\\bottomrule
    \end{tabular}
    \caption{\textbf{MNIST to MNIST-M} Ablation Results performed with pruning $40\%$ channels with either our proposed DSS score or the reverse of our DSS score, i.e. we remove channels with highest DSS instead. The effectiveness of DSS is demonstrated in boosting the generalization performance of the model. Notably, with our proposed DSS score, even before fine-tuning immediately following pruning, we could already observe improvement.}
    \label{tab:ablation}
\end{table}

\begin{table*}[t!]
    \centering
    \begin{adjustbox}{max width=\textwidth}
    \begin{tabular}{c|cc|c}
    \toprule
    Method & PACS & OfficeHome & Avg. \\ \toprule
    DANN~\cite{ganin2015unsupervised} & 83.6 & 65.9 & 74.8 \\
    CDANN~\cite{li2018domain} & 82.6 & 65.8 & 74.2 \\
    IRM~\cite{arjovsky2019invariant} & 83.5 & 64.3 & 73.9 \\
    GroupDRO~\cite{sagawadistributionally} & 84.4 & 66.0 & 75.2 \\
    ARM~\cite{zhang2021adaptive} & 85.1 & 64.8 & 75.0 \\
    ERM~\cite{vapnik1998statistical} & 84.2 & 67.6 & 75.9 \\
    Mixup~\cite{wang2020heterogeneous, xu2020adversarial, yan2020improve} & 84.6 & 68.1 & 76.4 \\
    SelfReg~\cite{kim2021selfreg} & 85.6 & 67.9 & 76.8 \\
    MIRO~\cite{cha2022domain} & 85.4 & 70.5 & 78.0 \\
    \rowcolor{lightgray}
    \textbf{MIRO + Ours $10\%$} & $\mathbf{86.5}$ & $\mathbf{71.4}$ & $\mathbf{79.0}$ \\
    \bottomrule
    \end{tabular}
    \end{adjustbox}
    \caption{\textbf{Domain Bed} Domain generalization results of applying our proposed method on state-of-the-art method MIRO. We reliably improve the performance of MIRO by shrinking the model size down by $10\%$ removing channels selected by our DSS score.}
    \label{tab:domainbed}
\end{table*}

\subsection{MNIST-M}

We conduct experiments on MNIST-M with a simple ConvNet which is also leveraged in works like ~\cite{ganin2015unsupervised}. Here, we train on the vanilla MNIST data and evaluate on MNIST-M data for generalization performance and robustness. The two important comparisons here are (1) baseline, model only trained on the source MNIST and tested on MNIST-M and (2) baseline pruned with standard pruning metric L2~\cite{conf/iclr/0022KDSG17}. We prune at various channel sparsity ratios to better understand the behaviors. Results and comparisons are summarized in Figure\ref{fig:mnist_res}. \par
We can first observe that, standard and existing simple pruning method L2~\cite{conf/iclr/0022KDSG17} can already offer a small improvement across various sparsity levels. This confirms our expectation that pruning can improve the generalization ability of the model.\par
Next, we can observe that our proposed DSS score further improves the results by a margin. Notably, we can improve the baseline performance by more than $5$ points by sparsifying $60\%$ of the channels in the model.\par
\subsubsection{Ablation}

Here, we also conduct a quick ablation study in verifying the effectiveness of our proposed method. Results are presented in Table \ref{tab:ablation}. We compare with pruning with the reverse of our metric, which means that instead of removing the most sensitive features with the smallest DSS score, we remove the most similar features with the highest DSS score. Expectedly, we observe that the method degrades the performance of the baseline model. Surprisingly, we observe that, with DSS, even before fine-tuning, we can observe an improvement over the baseline from $59.30$ to $61.14$, further strengthening the effectiveness of our score.

\section{DomainBed}
As promised, we could also combine our proposed method with any state-of-the-art algorithm and further boost the performance. The experiment here is conducted on datasets PACS and OfficeHome from DomainBed~\cite{gulrajanisearch}. We develop based on code provided by MIRO~\cite{cha2022domain} and follow the same suggested optimization settings. On PACS and OfficeHome, each contains four datasets, we follow the standard \emph{leave-one} procedure which performs training and evaluation four times with each time training on three distributions and testing on the other. Final score is then averaged over these four runs. Moreover, our DSS score is also computed with an average of all of the three training domains. Results are presented in Table.\ref{tab:domainbed}. We can further improve the state-of-the-art MIRO performance by introducing $10\%$ channel sparsity selected by our method. Compared with MIRO~\cite{cha2022domain}, on PACS, we improve it from $85.4$ to $86.5$; on OfficeHome, we improve it from $70.5$ to $71.4$. On average, we improve MIRO from $78.0$ to $79.0$, a whole point improvement over the state-of-the-art only by introducing sparsity into the model.
\section{Conclusion} In this paper, we study whether we could use pruning as a reliable method to improve the generalization performance of the model. We propose a novel pruning score DSS, designed not to maintain source accuracy as typical pruning work, but to directly enhance the robustness of the model. We conduct empirical experiments to validate our method and demonstrate that it can be even combined with state-of-the-art generalization work like MIRO\cite{cha2022domain} to further boost the performance.

\section{Acknowledgements} Thanks to Professor Tatsunori Hashimoto and his great course CS 329D at Stanford for inspiration to this paper.

\bibliography{main}
\bibliographystyle{icml2023}
67.199+ 56.930+ 77.872+ 82.014
\end{document}